\documentclass{article}

 \usepackage[preprint]{nips_2018}

\usepackage[utf8]{inputenc} % allow utf-8 input
\usepackage[T1]{fontenc}    % use 8-bit T1 fonts
\usepackage{hyperref}       % hyperlinks
\usepackage{url}            % simple URL typesetting
\usepackage{booktabs}       % professional-quality tables
\usepackage{amsfonts}       % blackboard math symbols
\usepackage{nicefrac}       % compact symbols for 1/2, etc.
\usepackage{microtype}      % microtypography
\usepackage{graphicx}
\usepackage{amsmath}
\usepackage{amssymb}
\usepackage{color}
\usepackage{enumitem}
\usepackage{multirow}
\usepackage{wrapfig}

\title{DNN or $k$-NN: \\ That is the Generalize vs. Memorize Question}

\author{
  Gilad Cohen and Raja Giryes\\
  School of Electrical Engineering\\
  Tel Aviv University\\
  Tel Aviv, 69978\\
  \texttt{\{giladco1@post, raja@tauex\}.tau.ac.il}\\
  \And
  Guillermo Sapiro\\
  Electrical and Computer Engineering\\
  Duke University\\
  North Carolina, 27707\\
  \texttt{guillermo.sapiro@duke.edu}
}

\begin{document}

\maketitle

\begin{abstract}
This paper studies the relationship between the classification performed by deep neural networks (DNNs) and the decision of various classical classifiers, namely $k$-nearest neighbours ($k$-NN), support vector machines (SVM) and logistic regression (LR), at various layers of the network. 
This comparison provides us with new insights as to the ability of neural networks to both memorize the training data and generalize to new data at the same time, where $k$-NN serves as the ideal estimator that perfectly memorizes the data. We show that memorization of non-generalizing networks happens only at the last layers. Moreover, the behavior of DNNs compared to the linear classifiers SVM and LR is quite the same on the training and test data regardless of whether the network generalizes. On the other hand, the similarity to $k$-NN holds only at the absence of overfitting. Our results suggests that $k$-NN behavior of the network on new data is a sign of generalization. Moreover, it shows that memorization and generalization, which are traditionally considered to be contradicting to each other, are compatible and complementary.
%The core question in this work is what is the representation learned by neural networks and 
%This simple important connection shown here provides a better understanding of the relationship between the ability of neural networks to generalize and their tendency to memorize the training data, which are traditionally considered to be contradicting to each other and here shown to be compatible and complementary.
\end{abstract}

\section{Introduction}
\label{Introduction}
Deep learning is considered as one of the strongest machine learning tools today, getting state-of-the-art results in many domains such as computer vision \cite{Krizhevsky2012ImageNetNetworks,Schroff2015FaceNet:Clustering,Voulodimos2018DeepReview}, natural language processing \cite{DBLP:journals/corr/BahdanauCB14,Kim2014ConvolutionalClassification}, and speech recognition \cite{Hinton2012DeepRecognition,DBLP:journals/corr/ZhangPBZLBC17}. The structure of the corresponding deep neural networks (DNNs) is very similar in all these applications and generalizes well in all of them on very large datasets.

The generalization error of a learning system is a measure of its ability to correctly predict new unseen data. Formally, the generalization error of a model is defined as the difference between the empirical error and the expected error, which is practically measured by the difference between the test and the training error. One of the important properties that DNNs demonstrate empirically is their ability to generalize well. 
%Nonetheless, it is still unclear why these networks demonstrate this unprecedented behavior and they are still mostly viewed as a black box.

Many theoretical attempts have been performed to explain the ability of neural networks to generalize well, e.g., \cite{arora2018stronger,articleb,2017arXiv171005468K,NIPS2017_7176,SokolicRobustLargeMargin}.
Yet, many of them consider the case where the number of training examples exceeds the number of parameters. In practice DNNs generalize well even in the over-parameterized case, where they have been shown to memorize the data \cite{Poggio2017TheoryOD,Zhang2017UnderstandingGeneralization}. 

Since memorization and generalization are classically considered to contradict each other, in this work we address the question of how these two can co-exist in DNNs. To study this, we focus on standard (state-of-the-art) networks used for image classification. We empirically show that DNNs exhibit a similar behavior to classical $k$-nearest neighbor ($k$-NN), Support Vector Machine (SVM), and Logistic Regression (LR) classifiers, applied on their learned feature space. We show that the DNN behaves similarly to the SVM and LR models for both the training and testing sets, regardless of the generalization of the network. Moreover, it is observed that the DNN behaves similarly to the $k$-NN only when the network generalizes well (no overfitting). The detailed study in this work supports the conjecture that DNN generalizes by learning a new metric space adapted to the structure of the given training data. Yet, at the same time, it memorizes this new feature space (and therefore the training data), and its predictions are based on a $k$-NN search with the metric in this new learned embedding space. This behavior of the network is maintained both for the training and testing sets, thereby {showing that DNNs both generalize and memorize. This is the first time that these previously considered contradictory DNN features, are shown to co-exists and even collaborate}.

We demonstrate the above behavior for CIFAR-10/100 and MNIST on three popular architectures: Wide-Resnet 28-10 \cite{DBLP:journals/corr/ZagoruykoK16}, LeNet \cite{LeCun1998Gradient-basedRecognition}, and a simple multi layer perceptron (MLP) composed of two fully connected layers. We compare the $k$-NN model prediction accuracy to the one of the DNN and show that the probability of the DNN to predict the exact same $k$-NN classification result per test sample increases as the number of training iterations increases and as the network deepens, approaching 100\% (full agreement between the $k$-NN and the DNN) for very deep networks at the end of the training.

\section{Related work}
\label{Related work}
Boiman et al. \cite{4587598} argued that $k$-NN classifiers should be applied on crafty spaces and not on the image space. They fitted a nearest neighbors algorithm on a set of image descriptors and outperformed other parametric classifiers. In addition, they showed that their algorithm can approximate the optimal Naive-Bayes classifier.

Papernot and McDaniel \cite{DBLP:journals/corr/abs-1803-04765} demonstrated that DNN classifier predictions are supported by NN predictions throughout the network. After training a DNN classifier with SGD using the cross-entropy loss, they applied a $k$-NN model on every hidden layer. Next, they calculated for every test input its softmax prediction and evaluated on what extent it is conformed with predictions from the $k$-NN models. They found that for unmodified test images the softmax label agrees with the majority of the $k$-NN predicted labels, but if adversarial noise is applied then the softmax label is not credible, lacking support from the training data. This work is the closest to ours, with the difference that we explicitly study the relationship between generalization and memorization in DNNs.

The work of Zhang et al. \cite{Zhang2017UnderstandingGeneralization} empirically demonstrated that DNNs have sufficient capacity to memorize entire datasets. This memorization is evident even when the dataset is not coherent; DNNs trained on randomized CIFAR-10 labels yielded perfect accuracies on the training set. These findings question how exactly DNNs are generalizing as they are shown to be memorizing even random data. %The authors demonstrate that adding regularization improves the generalization error only by a small margin, which cannot explain why DNNs are in practice very good generalizers inherently.

A following work hypothesized that stochastic gradient decent (SGD) optimizes to good solutions only if they are surrounded by a relatively large volume of solutions which are nearly as good \cite{articleb}. This work demonstrated nonvacuous bounds for a stochastic neural network on MNIST by minimizing the PAC-Bayesian bound \cite{McAllester1999PAC-BayesianAveraging} while perturbing the weights to capture flat minima.

Keskar et al. \cite{DBLP:journals/corr/KeskarMNST16} showed that the test accuracy usually degrades as one uses larger batch size in the training. They observed that large-batch methods tend to converge to sharp minima, which usually results in poor generalization. However, Dinh et al. \cite{DBLP:journals/corr/DinhPBB17} challenged this conjecture; they argue that good generalization can be maintained by altering the model parameters while decreasing the flatness of the loss function.

Smith and Le \cite{l.2018a.ABasianPerspective} showed that the perfect memorization phenomenon of randomly labeled datasets is not unique to deep networks, and is also replicated in a small over-parameterized linear model. They explain this observation by analyzing the Bayesian evidence that contains information about the curvature of the model's minima. Additionally, they revisited the results of Keskar el al. \cite{DBLP:journals/corr/KeskarMNST16}, finding that the mini-batch size plays a significant role in the generalization error. They showed empirically that there exist an optimal batch size and predicted its scaling relatively to the learning rate, training set size, and the optimizer's momentum coefficient.

While all the above works try to explain the generalization vs. memorization paradigm, here we empirically demonstrated for the first time that these two concepts are not contradictory but rather complementary and even collaborative.

\section{Model}

Many research groups have shown that DNNs are capable of memorizing large datasets \cite{Arpit2017ANetworks,Cheng2016WideSystems,Zhang2017UnderstandingGeneralization}. The most familiar and simple model that memorizes the data is $k$-NN, which assigns to every training sample a point in a feature space. This work provides an initial evidence that when the network generalizes, the  classification of the test data relies on their nearest neighbors from the training set.

We measure correspondence between the DNN softmax predictions to SVM, LR, and $k$-NN predictions applied on the various DNN layers, where the main focus is the embedding space; we use $k=30$ for all our experiments. To measure the distance between the decisions of each classifier we use the Kullback–Leibler (KL) divergence:
\begin{eqnarray}
\label{DKL}
D_{KL}(p||q) = \sum_i p_i \log(p_i/q_i), 
\end{eqnarray}
where $p_i$ and $q_i$ are the probabilities of selecting class $i$ with the fitted model ($k$-NN/SVM/LR) and the DNN respectively. Clearly, we calculate $D_{KL}(p||q)$ as the average for all training/testing samples. %By abuse of notation we use $D_{KL}(p|q)$ for the average also.
In addition, we introduce another measure for the consistency between a single DNN prediction to its corresponding $k$-NN/SVM/LR prediction. Let $S = \big\{(s_1,\textit{l}_1), ..., (s_N,\textit{l}_N)\big\}$, where $s_i$ is a given data point with its label $l_i$.
We define the probability $P_{SAME}$($M$) as a  measure of how close is the DNN softmax prediction to the one of a model $M$ ($k$-NN/SVM/LR):
\begin{eqnarray}
\label{$P_{SAME}$}
 \hspace{-0.3in} P_{SAME}(M) & \triangleq & p\big(\: \textit{f}_{M}(s) = \textit{f}_{DNN}(s) \:\big),
\end{eqnarray}
where 
$\textit{f}_{DNN}(s)$ and 
$\:\textit{f}_{M}(s)$ are the DNN and model  $M$  prediction functions for the pair $\big(s,\textit{l}\big)$, respectively.

\section{Experiments}
\begin{figure}[ht]
\includegraphics[width=\linewidth]{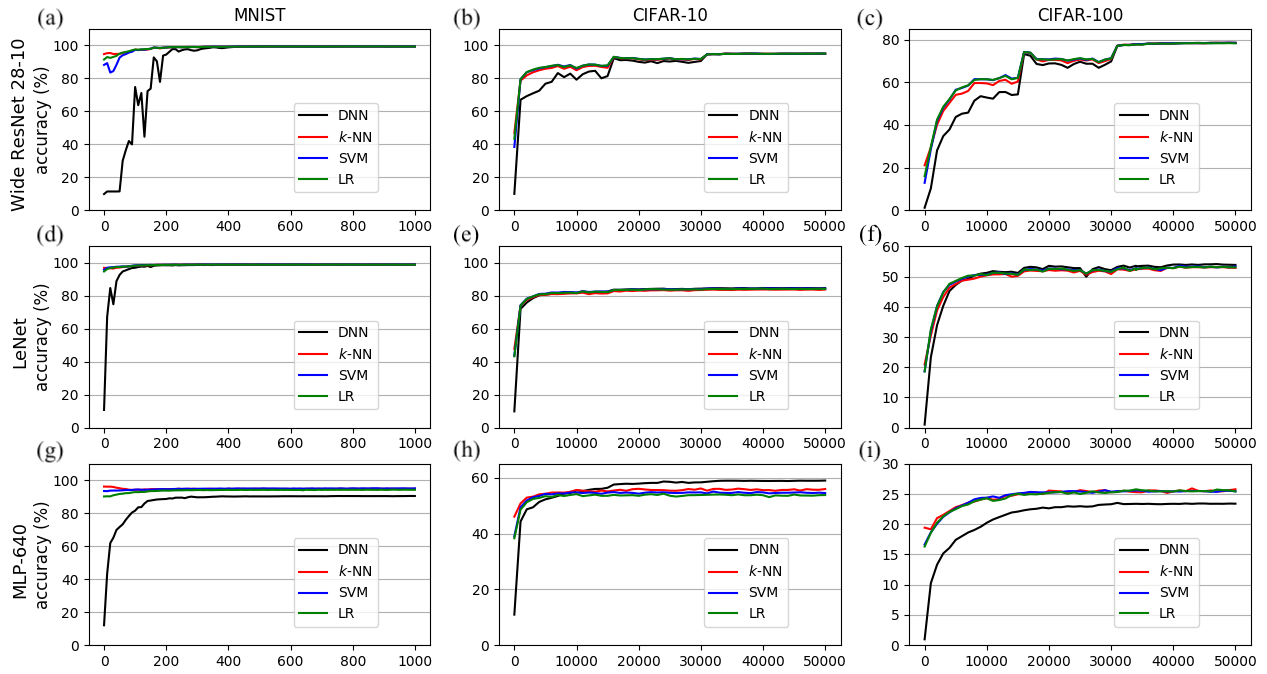}
\caption{Test accuracies as a function of training steps for DNN (black) and $k$-NN (red) with $k$=30, SVM (blue), and LR (green). Very high correspondence is observed between the DNN and predictions of the classic models, especially for the deeper networks.}
\label{images/test_acc_vs_iter}
\end{figure}

We checked the relationship between the DNN softmax to $k$-NN, SVM, and LR models fitted on the embedding space (before the last fully connected layer), by empirically evaluating three common network architectures: Wide-Resnet 28-10 \cite{DBLP:journals/corr/ZagoruykoK16}, LeNet \cite{LeCun1998Gradient-basedRecognition}, and a simple multilayer perceptron (MLP), denoted by MLP-640, which has two layers and a hidden layer of size $640$.
We evaluated these architectures on three popular datasets: MNIST, CIFAR-10 and CIFAR-100, and found that in all cases the LR, MLP and $k$-NN models yield the same test accuracy, performing as good as the DNN (Figure~\ref{images/test_acc_vs_iter}).

\begin{figure}[ht]
\includegraphics[width=\linewidth]{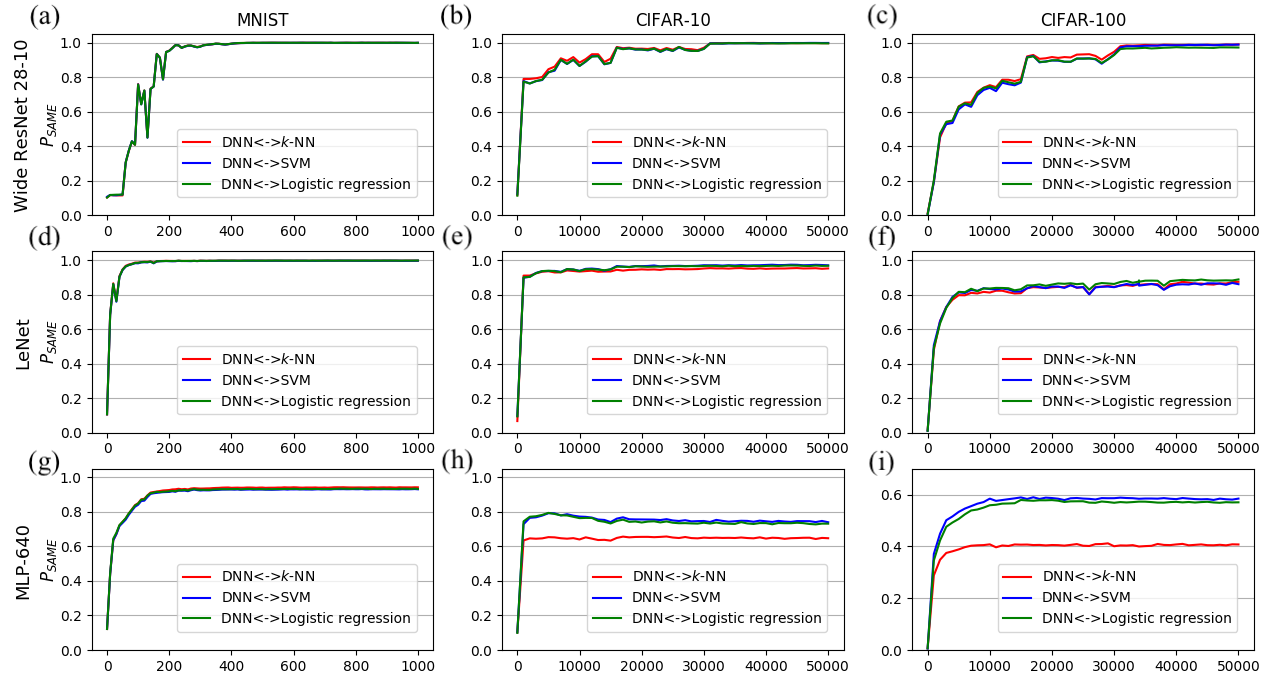}
\caption{$P_{SAME}$ value (Eq.~\eqref{$P_{SAME}$}) as a function of training steps for $k$-NN (red) with $k$=30, SVM (blue), and LR (green). Very high agreement between the models is observed, especially for the deeper networks.}
\label{images/test_psame_vs_iter}
\end{figure}

\begin{figure}[ht]
\includegraphics[width=\linewidth]{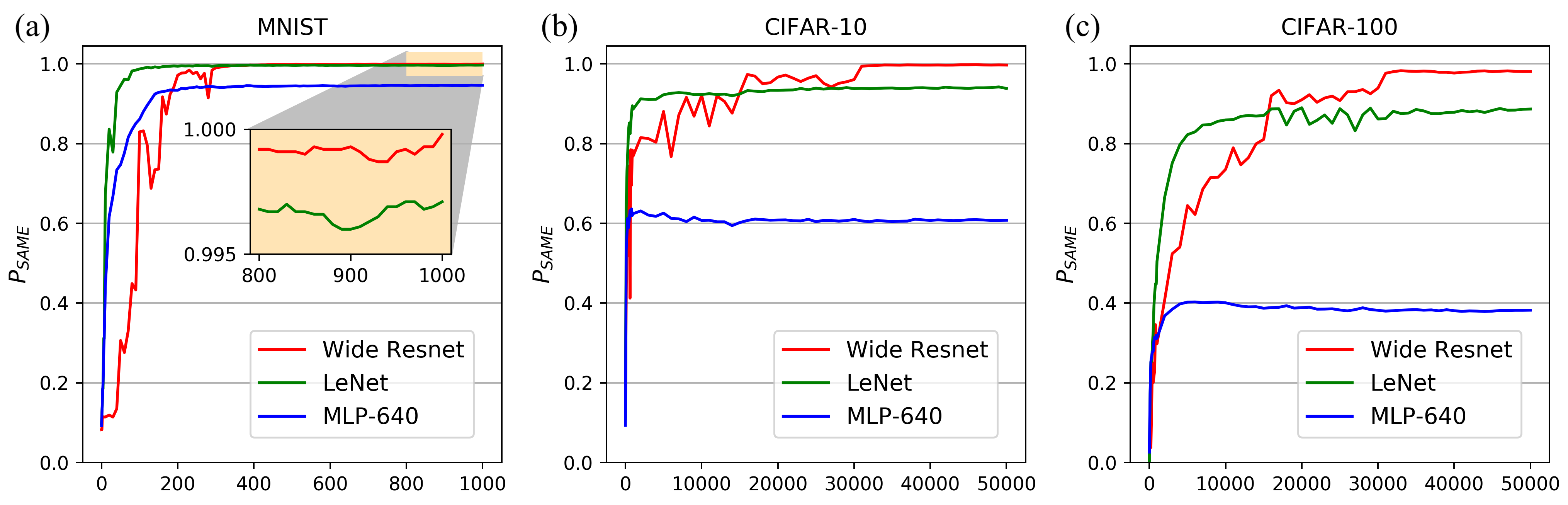}
% \vskip -0.01in
\caption{The $P_{SAME}$ measure between DNN and $k$-NN (described in Eq.~\eqref{$P_{SAME}$}) as a function of the training step. %(a), (b) and (c) correspond to MNIST, CIFAR-10 and CIFAR-100 datasets, respectively. 
The probability of the DNN and $k$-NN models to predict the same label increases during the training.}
%The deep Wide-Resnet architecture demonstrates very strong agreement between the models.}
\label{images/psame_scores_vs_iter_edit}
\end{figure}

Next, we calculated the $P_{SAME}$ value (Eq.~\eqref{$P_{SAME}$} between the DNN and the three classical models above for MNIST, CIFAR-10, and CIFAR-100 datasets (Figure~\ref{images/test_psame_vs_iter}). It is shown that the $k$-NN, SVM, and LR not only perform statistically as the DNN softmax, their predictions match the DNN softmax predictions with very high probability, notably for the deeper networks. The $k$-NN results in Figure~\ref{images/test_psame_vs_iter} are also portrayed in Figure~\ref{images/psame_scores_vs_iter_edit} with comparison between the different architectures. Observe that $P_{SAME}$ approaches $1$ for each dataset as the network deepens.
This demonstrates the similarity of DNN to the non-parametric $k$-NN model.

\begin{figure}
\centering
\includegraphics[width=\linewidth]{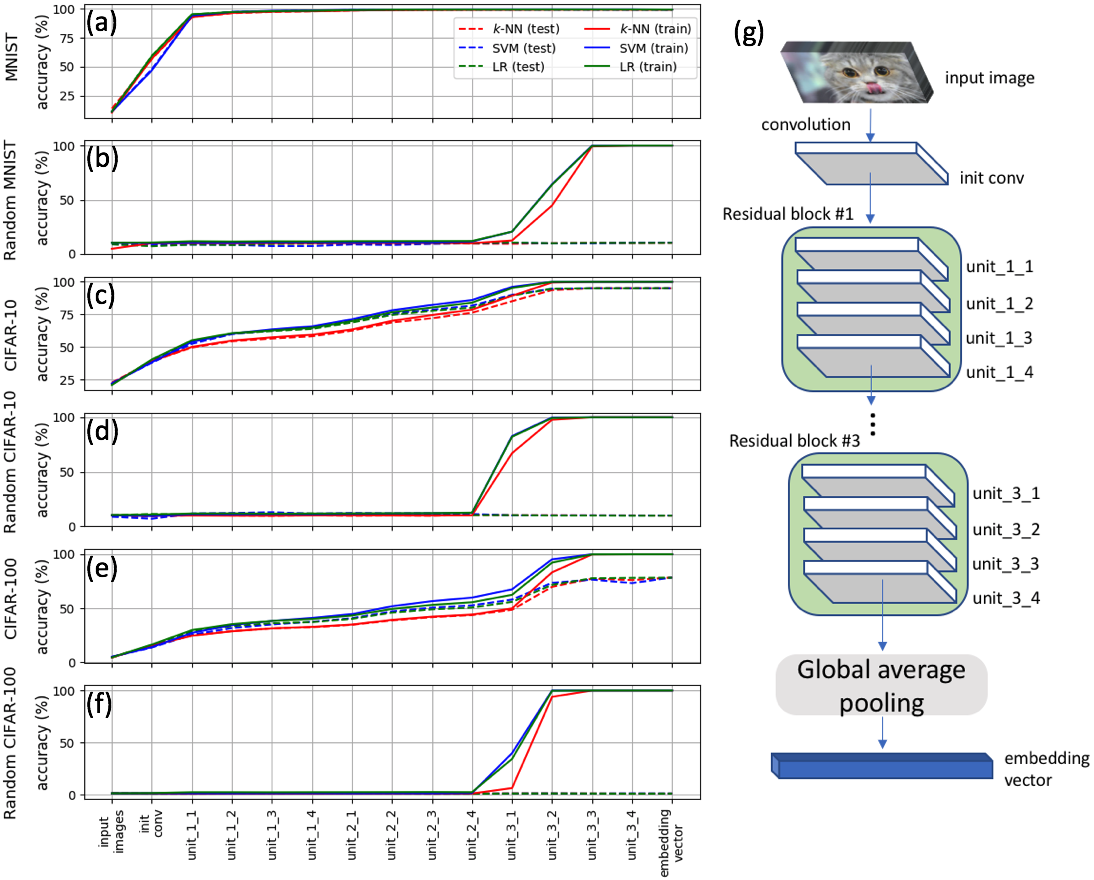}
\caption{Train and test accuracies as a function of the embedding layer depth for MNIST (a)$\&$(b), CIFAR-10 (c)$\&$(d), and CIFAR-100 (e)$\&$(f) and their random counterparts. After the training is complete, we fit $k$-NN, SVM, and LR models for every layer in the Wide-Resnet network (g) and use its embedding space to calculate the accuracy.}
\label{images/train_test_acc_vs_layer_wrn}
\end{figure}

Figures~\ref{images/train_test_acc_vs_layer_wrn}(a),(c),(e)) exhibit how well the DNN generalizes as a function of the layers depth by fitting the $k$-NN/SVM/LR models on every feature vector along the neural network layers. This demonstrate that DNNs' generalization improves gradually along the network layers. 
The layers' names are depicted in figure~\ref{images/train_test_acc_vs_layer_wrn}(g).
Figures~\ref{images/train_test_acc_vs_layer_wrn}(b),(d),(f) show the results of the same experiment but with random labels  (similarly to the setup in Zhang et al. \cite{Zhang2017UnderstandingGeneralization}). Observe that at the absence of generalization, the memorization occurs only within the last layers.

\begin{figure}[ht]
\centering
\includegraphics[width=\linewidth]{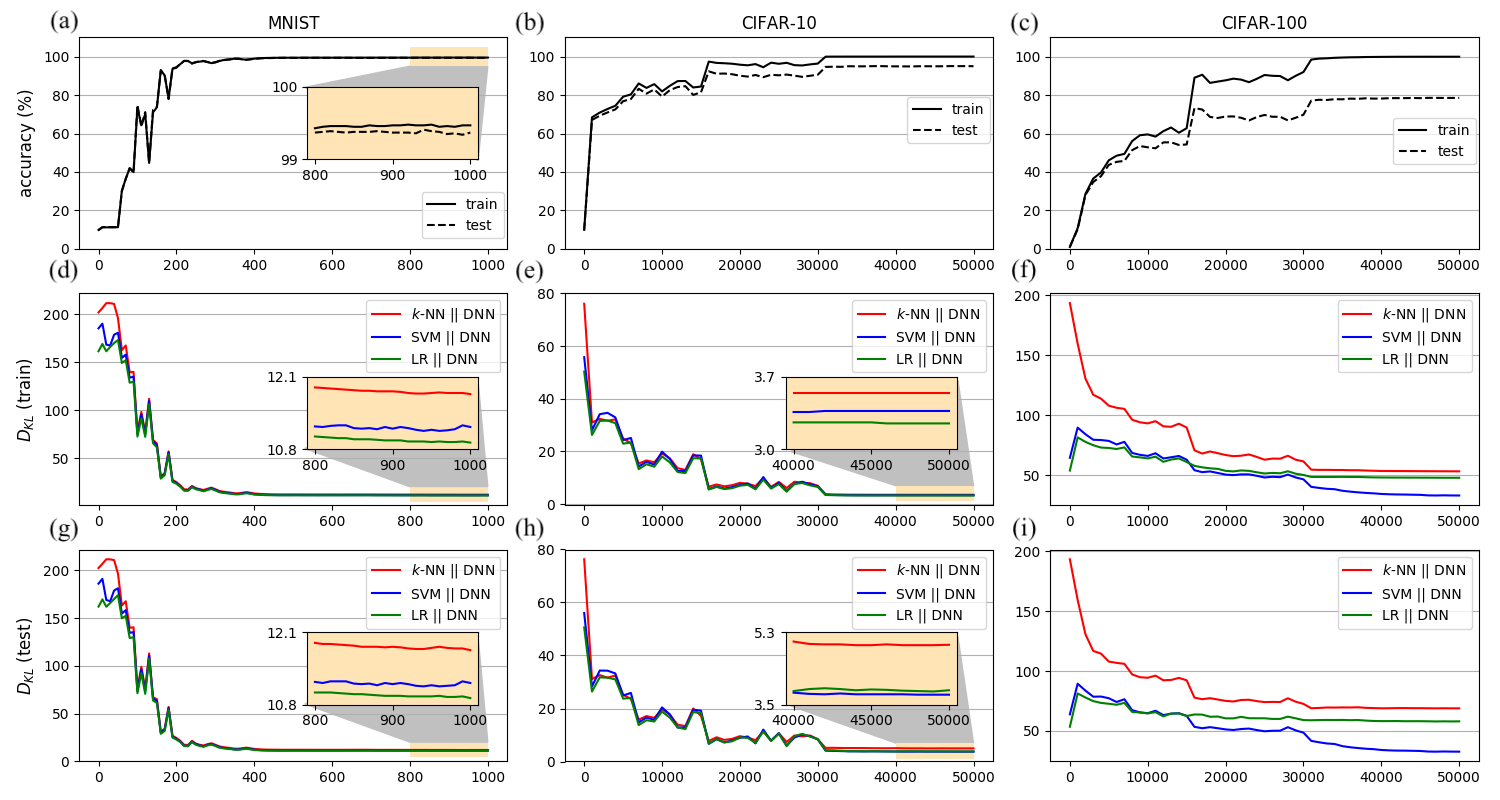}
\caption{$D_{KL}$ analysis of MNIST, CIFAR-10, and CIFAR-100 datasets. The top row shows the train and test accuracies. The middle and bottom rows show the $D_{KL}$ values (Eq.~\eqref{DKL}) of the training and testing sets, respectively, for $k$-NN/SVM/LR compared to the DNN softmax distribution. All values are presented as a function of the iteration step. The $D_{KL}$ trend of the testing set matches the training set throughout the training.}
\label{images/kl_div_trend}
\end{figure}

\begin{figure}[ht]
\centering
\includegraphics[width=\linewidth]{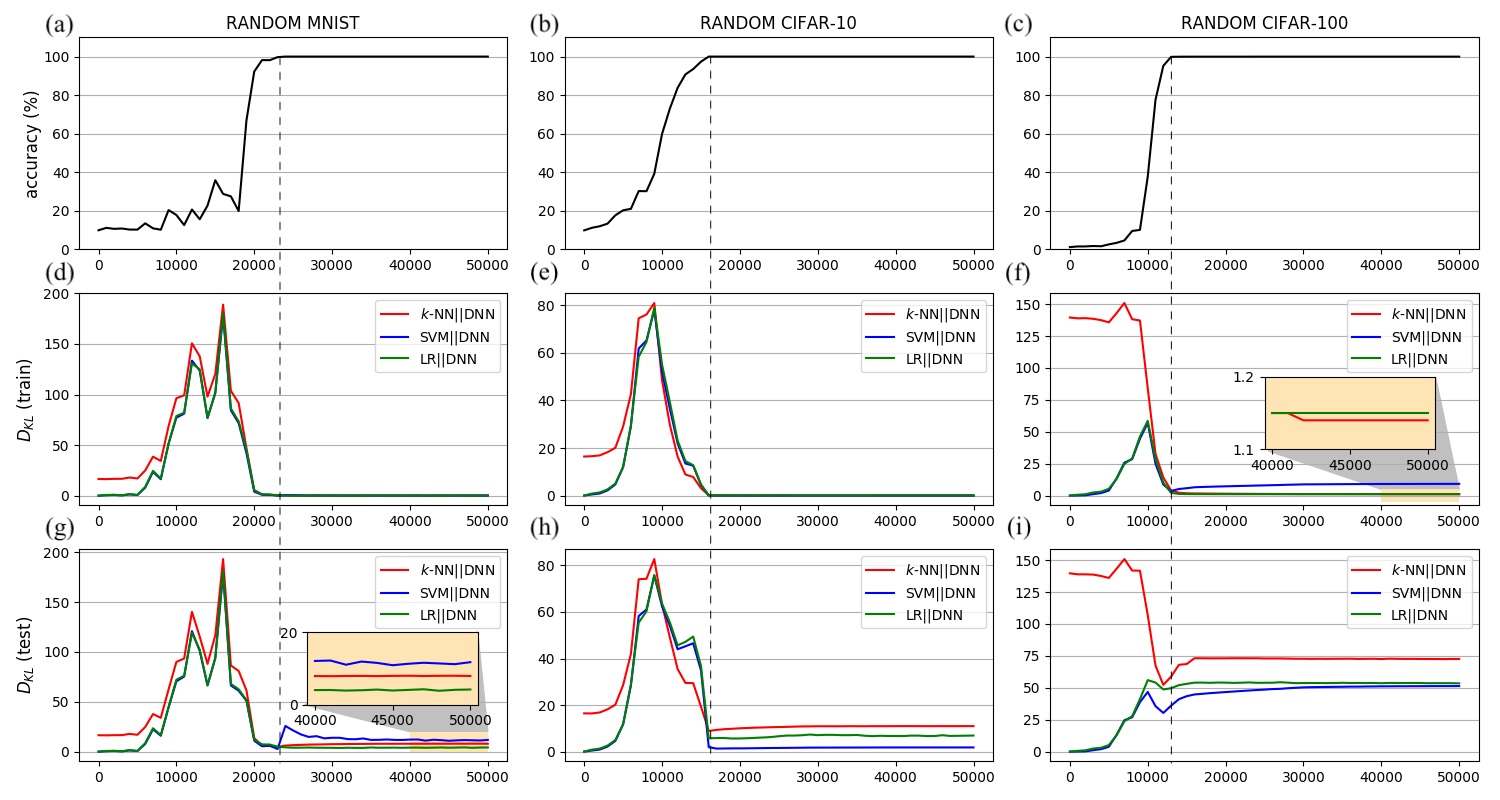}
\caption{$D_{KL}$ analysis of randomized datasets. The top row shows the train accuracies. The middle and bottom rows show the $D_{KL}$ values (Eq.~\eqref{DKL}) of the training and testing sets, respectively, for $k$-NN/SVM/LR compared to the DNN softmax distribution. All values are presented as a function of the iteration step. The vertical dashed lines mark the memorization of the training set, where the \textbf{train} accuracy reaches $100\%$. Contrary to the real (non randomized) case, the $D_{KL}$ trend of the testing set differs from the training set after the memorization occurs.}
\label{images/kl_div_trend_with_train_acc}
\end{figure}

Next, we demonstrate that when the Wide-Resnet is trained with real (non randomized) datasets, the $D_{KL}$($M$ || DNN) values (Eq.~\eqref{DKL}) of both the training set and testing set keep the same trend throughout the training, for every model $M$ ($k$-NN/SVM/LR) (Figure~\ref{images/kl_div_trend}). However, when the same experiment is repeated with randomized data, we observe that the KL divergence on the training set matches the KL divergence on the testing set just prior to the memorization (Figure~\ref{images/kl_div_trend_with_train_acc}). Memorization is defined as the moment where the \textbf{train} accuracy reaches $100.0\%$ for the first time, as depicted in the vertical dashed lines. Upon memorization the KL divergence trend between the training and testing sets dissipates. The training $D_{KL}$ values approach zero whereas the testing values increase to a constant value; this is notable mostly for $D_{KL}$($k$-NN || DNN) and less for SVM and LR (sometimes it happens and sometimes not). Thus, this sudden change of the $D_{KL}$($k$-NN || DNN) training and testing values can serve as an indicator for overfitting, where the network memorizes the training set excessively and weakens the generalization.

\begin{figure}[ht!]
\centering
\includegraphics[width=\linewidth]{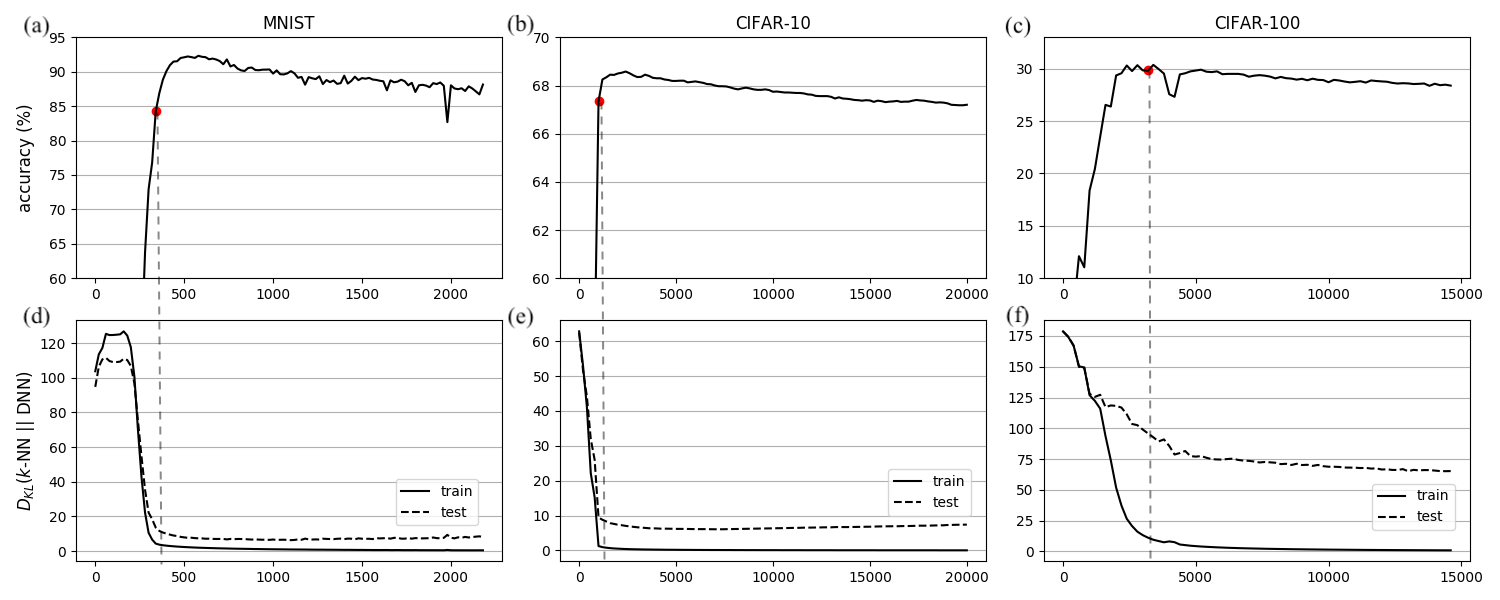}
\caption{$D_{KL}$ analysis in case of overfitting. The top row shows the test accuracies. The bottom row shows the $D_{KL}$($k$-NN || DNN) for both the training and testing sets. The vertical dashed lines mark the memorization of the training set, where the \textbf{train} accuracy reaches $100\%$. Prior to memorization, the $D_{KL}$($k$-NN || DNN) values resemble each other, however, after memorization they diverge.}
\label{images/kl_div_overfitting}
\end{figure}

To demonstrate this effect on real (non-randomized) datasets, we trained the Wide-Resnet network while enforcing overfitting (Figure~\ref{images/kl_div_overfitting}). This is achieved by using only a subset of the training set, without data augmentation and without weight decay regularization. The vertical dashed lines mark the memorization of the training set on the network.
One can observe that the $D_{KL}$($k$-NN || DNN) trend is maintained prior to the memorization for MNIST and CIFAR-10. After reaching memorization on the training set the consistency between the train and test values lessens. For CIFAR-100, this phenomena occurs slightly before the memorization, when the \textbf{train} accuracy equals $90\%$ (data not shown). In addition, for MNIST and CIFAR-10 the $D_{KL}$($k$-NN || DNN) test value rises slightly after the memorization. Yet, in all cases we find this correspondence between the memorization and the deviation between the trends of the KL-divergence in the test and train data.

\makeatletter
\setlength{\@fptop}{0pt}
\makeatother

\section{Discussion}
This work compared the behavior of DNN softmax to simple classifier model performed on its embedding space. All models yielded the same test accuracy, matching the DNN performance. In addition, we observed that the DNN softmax predictions matched the $k$-NN model with $\sim100\%$ probability (calculated using $P_{SAME}$ in Eq.~\eqref{$P_{SAME}$}) for deep networks; in other words, DNN classifiers approximate nearest neighbors decision on the embedding space. 

It has been also demonstrated that DNNs' generalization improves gradually along the network. We may claim that DNNs encapsulate the representation of the class labels within the dataset, generating better abstraction at each successive layer. While this is true for data with true labels, networks trained without any generalization (random lables), present memorization only at the last layers.  This observation may lead to new regularization techniques for training neural networks. % probably because the gradients are higher there. 

Lastly, we observed that when the network generalizes well, the KL divergence $D_{KL}$($k$-NN || DNN) maintains the same trend in the training and testing sets. However, when the network starts memorizing the training data on the expense of generalization (i.e. randomized data, overfitting), this value diverge, and usually rises in the testing set. This implies that DNNs approximates nearest neighbor decision when they generalize well.

\clearpage
\bibliography{nips_2018.bib}
\bibliographystyle{abbrvnat}
\small

\end{document}